\documentclass[10pt,twocolumn,letterpaper]{article}

\usepackage[pagenumbers]{iccv} 

\definecolor{iccvblue}{rgb}{0.21,0.49,0.74}
\usepackage[pagebackref,breaklinks,colorlinks,allcolors=iccvblue]{hyperref}
\usepackage{multirow}

\def\paperID{15614} 
\def\confName{ICCV}
\def\confYear{2025}

\title{EDM: Efficient Deep Feature Matching}
\author{
Xi Li \qquad Tong Rao \qquad Cihui Pan \\
Realsee \\
{\tt\small \{lixi042, raotong001, pancihui001\}@realsee.com}
}

\begin{document}
\maketitle
\begin{abstract}
    Recent feature matching methods have achieved remarkable performance but lack efficiency consideration.
    In this paper, we revisit the mainstream detector-free matching pipeline and improve all its stages considering both accuracy and efficiency.
    We propose an Efficient Deep feature Matching network, EDM.
    We first adopt a deeper CNN with fewer dimensions to extract multi-level features.
    Then we present a Correlation Injection Module that conducts feature transformation on high-level deep features,
    and progressively injects feature correlations from global to local for efficient multi-scale feature aggregation,
    improving both speed and performance.
    In the refinement stage,
    a novel lightweight bidirectional axis-based regression head is designed to directly predict subpixel-level correspondences from latent features,
    avoiding the significant computational cost of explicitly locating keypoints on high-resolution local feature heatmaps.
    Moreover, effective selection strategies are introduced to enhance matching accuracy.
    Extensive experiments show that our EDM achieves competitive matching accuracy on various benchmarks and exhibits excellent efficiency,
    offering valuable best practices for real-world applications.
    The code is available at \url{https://github.com/chicleee/EDM}.
\end{abstract}
\section{Introduction}
\label{sec:intro}
\begin{figure}[t]
      \centering
      \includegraphics[width=1\linewidth]{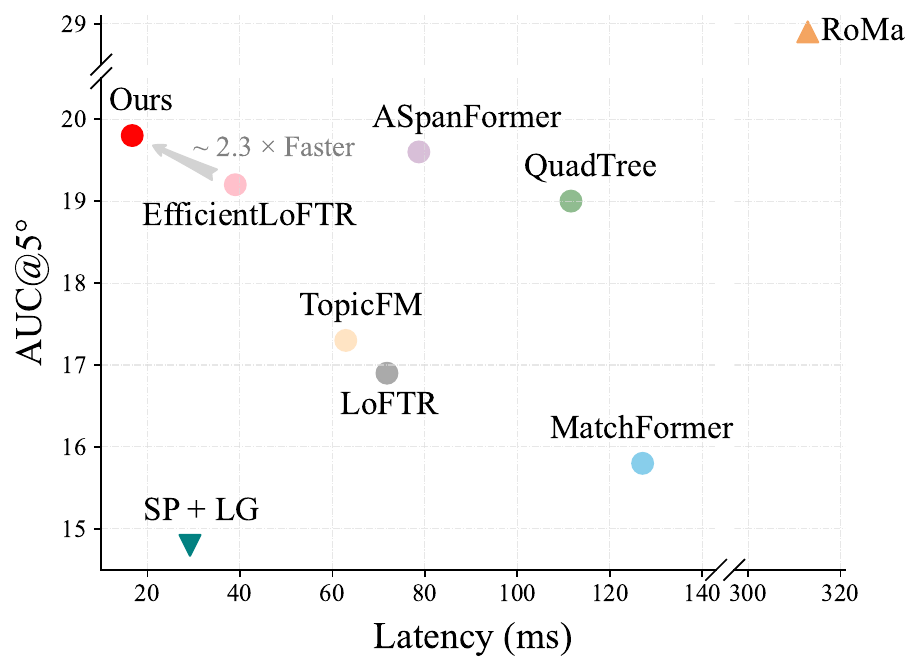}

      \caption{
            \textbf{Comparison of Matching Accuracy and Latency.}
            Our method achieves competitive accuracy with lower latency.
            Models are evaluated on the ScanNet dataset to get AUC@$5^{\circ}$ accuracy,
            while the latency for an image pair with 640$\times$480 resolution is measured on a single NVIDIA 3090 GPU.
      }
      \label{fig:performance}
\end{figure}
Image feature matching is a crucial task in the field of computer vision with a broad range of important applications,
including structure from motion (SfM)\cite{schonberger2016structure, agarwal2009building, he2024detector},
simultaneous localization and mapping (SLAM) \cite{mur2015orb, campos2021orb},
visual tracking \cite{shi1994good, zhao2008differential}, and visual localization \cite{sarlin2019coarse, sarlin2021back}, etc.
Traditional feature matching methods generally consist of several stages,
including keypoint detection, feature description and matching\cite{lowe2004distinctive, rublee2011orb, bay2006surf, calonder2011brief}.

Benefiting from the powerful feature description capability of deep neural networks,
many recent studies \cite{detone2018superpoint, revaud2019r2d2, barroso2019key, zhao2022alike} have adopted convolutional neural networks to extract local features,
which significantly outperform the conventional handcrafted features.
Besides, feature matching methods \cite{sarlin2020superglue, lindenberger2023lightglue} based on deep learning have also emerged and achieved remarkable results.
Although these methods are generally effective,
they still encounter difficulties due to various challenging factors, including illumination variations, scale changes, poor textures, and repetitive patterns.

To address these limitations, end-to-end detector-free local feature matching methods are coming into existence.
Early methods \cite{rocco2018neighbourhood, rocco2020efficient, li2020dual, zhou2021patch2pix, efe2021dfm} typically used the cost volume and neighborhood consensus to generate matches.
Given the powerful capability of modelling long-range global context information,
some studies \cite{sun2021loftr, wang2024eloftr, chen2022aspanformer} have started using Transformer \cite{vaswani2017attention} to establish precise correspondences.
To reduce computational complexity, most of these methods usually adopt a coarse-to-fine scheme.
Specifically, coarse matches at the patch level are first obtained using the nearest neighbor criterion, then refined to the sub-pixel level for increased accuracy.
More recently, some studies \cite{edstedt2023dkm, edstedt2024roma} have explored methods for generating dense, pixel-wise matching, achieving impressive performance on mainstream datasets.

Although previous methods have constantly achieved breakthroughs in matching accuracy,
few studies have focused on the ease of deployment and inference efficiency,
which limits their application in real-time programs.
Local feature matching is considered as a low-level computer vision task.
Consequently, the current mainstream methods prioritize high-resolution local features for accurate matching,
and their networks are designed to be typically shallow and wide,
resulting in limited utilization of global high-level contextual information.
While high-resolution local features offer superior localization accuracy and intuitive understanding,
they come at a significant computational cost.
A key insight is that focusing excessively on local details is computationally burdensome and superfluous.

In this work,
we introduce EDM,
an innovative and efficient deep feature matching network.
By extracting high-level feature correlations between two images at deeper layers and implicitly estimating precise fine matches,
EDM strikes an optimal balance between efficiency and performance.
\cref{fig:performance} highlights the impressive results of EDM.

In summary, the contributions of this paper include:
\begin{itemize}
      \item A new detector-free matcher significantly improves efficiency while maintaining competitive accuracy by redesigning all the steps of the mainstream paradigm.
      \item A Correlation Injection Module models deep features correlations with high-level context information
            and integrates global and local features by hierarchical correlation injection to enhance performance and efficiency.
      \item A novel lightweight bidirectional axis-based regression head for estimating subpixel-level matches implicitly.
      \item Efficient matching selection strategies are proposed to improve accuracy for both coarse and fine stages.
\end{itemize}
\section{Related Work}
\label{sec:relatedwork}
\begin{figure*}
	\centering
	\includegraphics[width=1\linewidth]{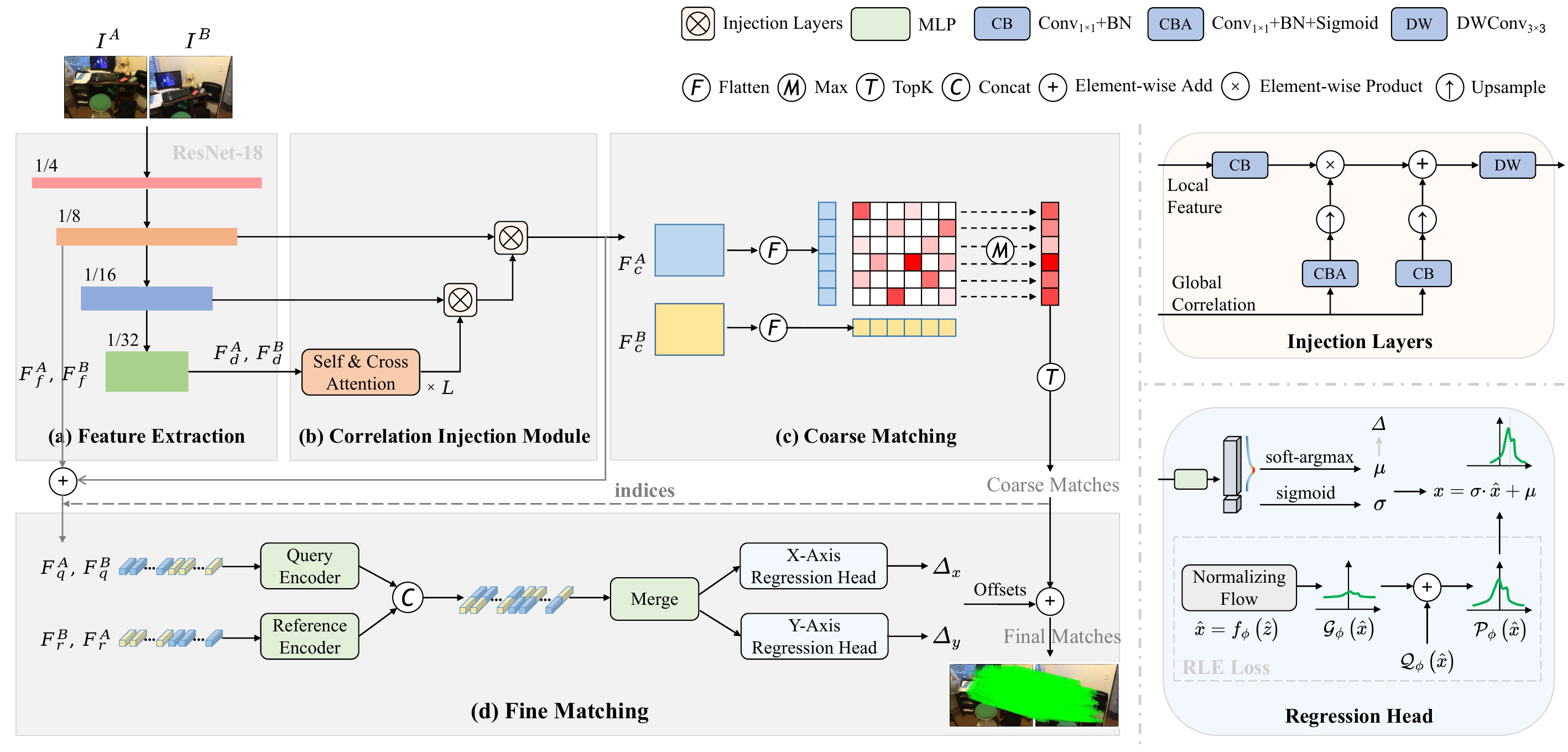}
	\caption{\textbf{Pipeline Overview}.
		\textbf{(a)} A deeper CNN backbone is adopted to extract multi-level feature maps.
		\textbf{(b)} In the Correlation Injection Module, we alternately apply self-attention and cross-attention a total of $L$ times to capture and transform the correlations between deep feature $F_{d}^{A}$ and $F_{d}^{B}$.
		Subsequently, two Injection Layers are employed to progressively inject feature correlations from deep to local levels.
		\textbf{(c)} After the CIM, the coarse features ${F}_{c}^{A}$ and ${F}_{c}^{B}$ are flattened and then correlated to produce the similarity matrix.
		To establish coarse matches, we determine the row-wise maxima in the probability matrix and select the top $K$ values among them.
		\textbf{(d)} For fine-level matching, the corresponding fine features are extracted by the indices obtained from the coarse matching process.
		We treat the fine features $F_{q}^{A}$ and $F_{q}^{B}$ as queries,
		while considering the same features but in reversed order, $F_{r}^{B}$ and $F_{r}^{A}$, as references.
		The query and reference features are encoded separately and then merged together.
		Then, a lightweight regression head is designed to estimate the reference offsets on the X and Y axes, respectively.
		The final matches are obtained by adding the coarse matches to their corresponding offsets.
	}
	\label{fig:network}
\end{figure*}
\subsection{Feature Matching}
\noindent\textbf{Sparse Matching.}
Sparse matching methods are also known as detector-based methods.
Classical methods utilize handcrafted keypoint detection, feature description and matching \cite{lowe2004distinctive,bay2006surf, rublee2011orb, calonder2011brief}.
Recently, learning-based keypoint detection \cite{ rosten2006machine, barroso2019key}
, description \cite{yi2016lift, zhao2022alike, gleize2023silk, potje2024xfeat}
and matching methods \cite{chen2021learning, shi2022clustergnn, jiang2024omniglue, kim2024keypt2subpx}
leverage the powerful expressive capabilities of deep neural networks to enhance their robustness and performance.
Notably, SuperPoint (SP) \cite{detone2018superpoint} introduces a self-supervised network for both detection and description by leveraging homographic adaptation.
Numerous subsequent methods \cite{dusmanu2019d2, revaud2019r2d2, tyszkiewicz2020disk, luo2020aslfeat} follow this paradigm.
SuperGlue (SG) \cite{sarlin2020superglue} is the first to introduce the self- and cross-attention \cite{vaswani2017attention} to capture keypoint feature correlations, resulting in improved matching accuracy.
To improve efficiency, LightGlue (LG) \cite{lindenberger2023lightglue} finds that the computationally complex attention process can end earlier for most easy image pairs.
For sparse methods, detecting repeatable keypoints is still challenging, particularly in low-texture areas.
\\\noindent\textbf{Dense Matching.}
Dense matching methods aim to estimate all matchable pixel-level correspondences.
Earlier methods NCNet \cite{rocco2018neighbourhood} and its subsequent works \cite{rocco2020efficient, li2020dual} achieved end-to-end dense matching
by using 4D cost volume to represent features and possible matches.
More recently, DKM \cite{edstedt2023dkm} models the dense matches as probability functions with the Gaussian process and achieves impressive results.
Similarly, RoMa \cite{edstedt2024roma} is a dense matcher that leverages a frozen pre-trained DINOv2 \cite{oquab2023dinov2} model for extracting coarse features and a specialized VGG \cite{simonyan2014very} model for further refinement.
Dense matching methods exhibit significant matching capabilities,
but they tend to be slower in practical applications due to excessive computational overhead.
\\\noindent\textbf{Semi-Dense Matching.}
Semi-dense matchers\cite{zhou2021patch2pix, efe2021dfm} adopt a coarse-to-fine manner,
which not only fully utilizes the entire image space, but also avoids overly dense pixel-level calculations.
Benefiting from the long-range modeling capability of the Transformer \cite{vaswani2017attention},
LoFTR \cite{sun2021loftr} and its follow-ups \cite{chen2022aspanformer, tang2022quadtree,wang2022matchformer} apply the Transformer to enhance local features.
TopicFM \cite{giang2023topicfm} attempts to model high-level contexts and latent semantic information as topics in deeper layer features,
but it still uses the heavy fine-level network of LoFTR \cite{sun2021loftr}.
EfficientLoFTR (ELoFTR) \cite{wang2024eloftr} introduces an aggregated attention network to reduce local feature tokens for efficient transformation
and a correlation refinement module for fine-level correspondences location in high-resolution features,
achieves comparable performance with lower latency.
ETO \cite{ni2024eto} introduces multiple homography hypotheses for local feature matching, achieves comparable efficiency but displays a performance gap.
\subsection{Keypoints Estimation in Related Tasks}
Keypoints estimation is an important component of feature matching and also plays a significant role in various other computer vision tasks,
such as object detection \cite{duan2019centernet, zhou2019objects}, human pose estimation \cite{nibali2018numerical, li2021human,li2022simcc, lu2024rtmo}, hand and facial keypoints detection \cite{chen2022mobrecon, jin2021pixel}, etc.
DSNT \cite{nibali2018numerical} introduces the soft-argmax method to calculate the approximate maximum response point from the feature maps,
enabling the model to directly regress the coordinate values.
RLE \cite{li2021human} proposes an effective regression paradigm,
namely residual log-likelihood estimation,
which improves regression performance by utilizing normalized flows \cite{dinh2016density} to estimate latent distributions and facilitate the training process.
SimCC \cite{li2022simcc} divides each pixel into several bins and classifies the coordinates of each region to achieve subpixel-level positioning accuracy.
We design our fine matching network based on these methods to avoid the heavy computational burden of upsampling and high-resolution heatmaps.
\section{Methods}
\label{sec:methods}
An overview of our pipeline is shown in \cref{{fig:network}}.
\subsection{Feature Extraction}
Unlike previous detector-free methods \cite{sun2021loftr,wang2024eloftr} using a shallow-wide CNN to extract features at $\frac{1}{8}$ scale of the original image resolution for feature transformation and coarse-level matching,
and then employing the Feature Pyramid Network (FPN) \cite{lin2017feature} to upsample the features to $\frac{1}{2}$ or a higher scale for fine-level matching,
our feature extractor is a ResNet-18 \cite{he2016deep} with fewer channels and deeper layers.
In order to achieve higher efficiency and capture more comprehensive high-level context information such as semantics and geometries,
we utilize low-resolution deep feature maps $F_{d}^{A}$ and $F_{d}^{B}$ at $\frac{1}{32}$ scale for feature transformation and $F_{f}^{A}$ and $F_{f}^{B}$ at $\frac{1}{8}$ scale for fine matching regression.
\subsection{Correlation Injection Module}
Inspired by \cite{zhang2022topformer,lin2017feature}, the Correlation Injection Module (CIM) is introduced to aggregate the multi-scale features before coarse matching.
The CIM is composed of stacked Transformers and two Injection Layers (ILs) as a whole.
\\\noindent\textbf{Feature Transform.}
The deep feature maps $F_{d}^{A}$ and $F_{d}^{B}$ at $\frac{1}{32}$ scale are transformed by interleaving self- and cross-attention $L$ times
to obtain the correlations between the features of two images.
This design significantly reduces the token sequence length and computational overhead in the Transformer.
Following \cite{wang2024eloftr}, the 2D rotary positional embedding (RoPE) \cite{su2021roformer} is inserted to each self-attention layers to capture the relative spatial information.
\\\noindent\textbf{Query-Key Normalized Attention.}
Attention mechanism is a core component in the Transformer,
characterized by query ${Q}$, key ${K}$, and value ${V}$.
The attention weight, determined by ${Q}$ and ${K}$, results in an output that is a weighted sum of ${V}$.
To enhance the correlation modeling capability, we replace the vanilla attention \cite{vaswani2017attention} with Query-Key Normalized Attention (QKNA) \cite{henry2020query}, which is defined as:
\begin{equation}
	QKNormAtt(Q,K,V)=softmax(s \cdot \hat{Q}\hat{K}^{T})V
	\label{eq:cos_attention}
\end{equation}
where $s$ is a manual scale factor,
$\hat{Q}$ and $\hat{K}$ are obtained by applying L2 normalization in the head dimensions.
\\\noindent\textbf{Injection Layers.}
After modeling feature correlations, two cascaded Injection Layers (ILs) are used to upsample features to $\frac{1}{8}$ scale.
As illustrated in the top-right of the \cref*{fig:network},
the ILs take the backbone local features and the deep features containing global correlations as inputs.
The local features are passed through a 1$\times$1 convolution layer and a batch normalization layer in sequence (CB) to increase the number of channels to match the global features.
The low-resolution deep features,
which have a larger receptive field and contain global correlations and rich context information,
are first fed into the 1$\times$1 convolution, batch normalization and a sigmoid activation function (CBA) to generate weights to determine how much detail to retain for the local features.
Then, the output is upsampled to match the size of the local features and injected into the local features by element-wise product.
Meanwhile, the global features are passed through another CB block and bilinear interpolation upsampling,
and then element-wise added to the features after injection.
Additionally, a 3$\times$3 depthwise convolution (DW) \cite{howard2017mobilenets} is used to alleviate the aliasing effect of upsampling.
Finally, after two consecutive ILs, the multi-scale features from two views are efficiently aggregated,
and coarse features $F_{c}^{A}$ and $F_{c}^{B}$ for the subsequent matching process are obtained.
\begin{table*}[htbp]
	\footnotesize
	\centering
	\begin{tabular}{llccccccc}
		\toprule
		\multirow{2}*{Category} & \multirow{2}*{Method}                                                & \multicolumn{3}{c}{MegaDepth} & \multicolumn{3}{c}{ScanNet} & \multirow{2}*{Time (ms)}                                                                         \\
		\cmidrule(lr){3-5}
		\cmidrule(lr){6-8}
		                        &                                                                      & AUC@$5^{\circ}$               & AUC@$10^{\circ}$            & AUC@$20^{\circ}$         & AUC@$5^{\circ}$ & AUC@$10^{\circ}$ & AUC@$20^{\circ}$ &

		\\\midrule
		\multirow{2}*{Sparse}
		                        & SP \cite{detone2018superpoint} + SG \cite{sarlin2020superglue}       & 49.7                          & 67.1                        & 80.6                     & 16.2            & 32.8             & 49.7             & 48.4          \\
		                        & SP \cite{detone2018superpoint} + LG \cite{lindenberger2023lightglue} & 49.9                          & 67.0                        & 80.1                     & 14.8            & 30.8             & 47.5             & 29.2
		\\\midrule
		\multirow{2}*{Dense}
		                        & DKM \cite{edstedt2023dkm}                                            & 60.4                          & 74.9                        & 85.1                     & 26.6            & 47.1             & 64.2             & 186.2         \\
		                        & ROMA \cite{edstedt2024roma}                                          & 62.6                          & 76.7                        & 86.3                     & 28.9            & 50.4             & 68.3             & 312.9
		\\\midrule
		\multirow{7}*{Semi-Dense}
		                        & LoFTR \cite{sun2021loftr}                                            & 52.8                          & 69.2                        & 81.2                     & 16.9            & 33.6             & 50.6             & 71.8          \\
		                        & QuadTree \cite{tang2022quadtree}                                     & 54.6                          & 70.5                        & 82.2                     & 19.0            & 37.3             & 53.5             & 111.6         \\
		                        & MatchFormer \cite{wang2022matchformer}                               & 53.3                          & 69.7                        & 81.8                     & 15.8            & 32.0             & 48.0             & 127.1         \\
		                        & ASpanFormer \cite{chen2022aspanformer}                               & 55.3                          & 71.5                        & 83.1                     & 19.6            & \textbf{37.7}    & \textbf{54.4}    & 78.7          \\
		                        & TopicFM \cite{giang2023topicfm}                                      & 54.1                          & 70.1                        & 81.6                     & 17.3            & 35.5             & 50.9             & 62.9          \\
		                        & EfficientLoFTR \cite{wang2024eloftr}                                 & 56.4                          & 72.2                        & 83.5                     & 19.2            & 37.0             & 53.6             & 39.0          \\
		                        & Ours                                                                 & \textbf{57.5}                 & \textbf{73.2}               & \textbf{84.2}            & \textbf{19.8}   & 37.5             & \textbf{54.4}    & \textbf{16.7} \\
		\bottomrule
	\end{tabular}
	\caption{\textbf{Results of Relative Pose Estimation on the MegaDepth Dataset and ScanNet Dataset.}
		The models are trained on the MegaDepth dataset to evaluate all methods on both datasets.
		The AUC of relative pose error at multiple thresholds, and the average inference time on the ScanNet dataset for pairwise image of 640$\times$480 resolution is provided.
	}
	\label{tab:megadepth_and_scannet}
\end{table*}
\subsection{Coarse Matching}
We establish coarse-level matches based on the coarse feature maps ${F}_{c}^{A}$ and ${F}_{c}^{B}$ after correlation injection.
Each pixel on the feature maps ${F}_{c}^{A}$ and ${F}_{c}^{B}$ represents an 8$\times$8 grid region in original images.
So coarse matches indicate rough local window correspondences between two images.
Firstly, the coarse feature maps ${F}_{c}^{A}$ and ${F}_{c}^{B}$ are flattened to 1-D vectors $ {\tilde{F}}_{c}^{A}$ and ${\tilde{F}}_{c}^{B}$.
Then we utilize the inner product to build a similarity matrix $\mathcal{S}$ as follows:
\begin{equation}
	\mathcal{S} = \frac{\left \langle {\tilde{F}}_{c}^{A}(i), {\tilde{F}}_{c}^{B}(j)\right \rangle}{\tau}
	\label{eq:similarity}
\end{equation}
where ${\tau}$ means the temperature parameter.

Following \cite{sun2021loftr}, the matching probability matrix $\mathcal{P}_{c}$ is obtained by a dual-softmax \cite{rocco2018neighbourhood} operator on both dimensions of $\mathcal{S}$:
\begin{equation}
	\mathcal{P}_{c} = softmax(\mathcal{S}(i,\cdots))_{j} \cdot softmax(\mathcal{S}(\cdots,j))_{i}
	\label{eq:probability}
\end{equation}
\\\noindent\textbf{Efficient Implementation.}
We note that the above \cref{eq:probability} can also be implemented by
first calculating the exponential function as $\mathcal{Z} = e^{\mathcal{S}}$ only once,
and then taking the product of its row-wise and column-wise L1 normalizations,
so as to reduce redundant computations and improve inference efficiency.
This implementation can be defined as:
\begin{equation}
	\mathcal{P}_{c} = \frac{\mathcal{Z}}{\left \|\mathcal{Z}(i,\cdots)_{j} \right \|_{1}}\cdot \frac{\mathcal{Z}}{\left \|\mathcal{Z}(\cdots,j)_{i} \right \|_{1}}
	\label{eq:probability_opt}
\end{equation}
\\\noindent\textbf{Coarse Matching Selection.}
Contrary to the previous methods of selecting matches using Mutual Nearest Neighbor (MNN),
we first obtain the maximum values from each row of the probability matrix $\mathcal{P}_{c}$,
and then select the Top-K scoring values to control the number of coarse matches.
Besides, the selected matching probabilities should be higher than the coarse-level threshold $\theta _{c}$.
Such a matching selection strategy drastically reduces the time complexity,
and the elimination of dynamic tensor shapes facilitates the formation of mini-batches for efficient inference.
\subsection{Fine Matching}
\begin{figure}[t]
	\centering
	\includegraphics[width=1\linewidth]{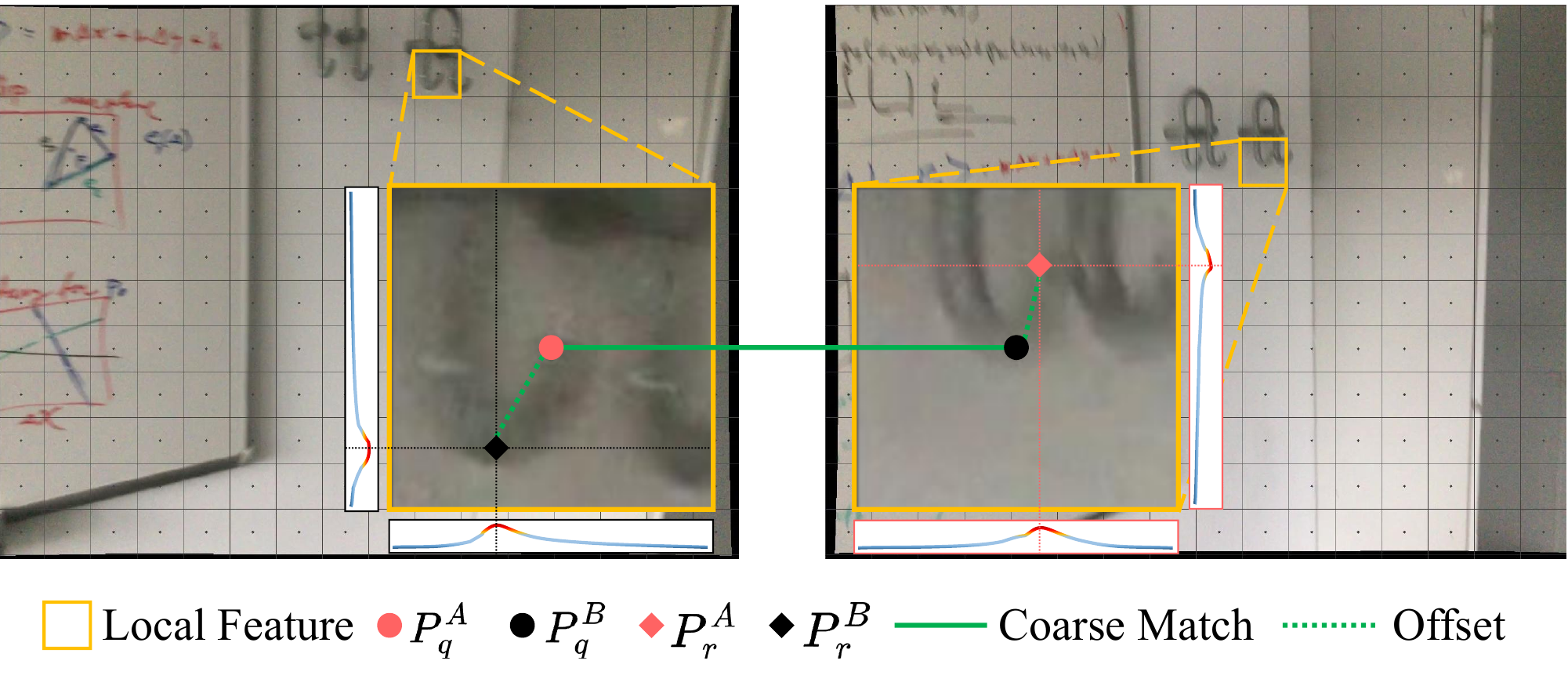}
	\caption{\textbf{Bidirectional Refinement.}
		For a coarse matching pair, the center point of one grid serves as query for fine matching, and its corresponding reference point is offset from the center point in another grid, exhibiting duality.}
	\label{fig:grid}
\end{figure}
For higher efficiency, we regress fine-level matching offsets directly from latent features, abandoning explicit pixel-level keypoint localization from high-resolution features.
Firstly, we take the element-wise sum of backbone features $F_{f}^{A}$ , $F_{f}^{B}$ and coarse-level features ${F}_{c}^{A}$ , ${F}_{c}^{B}$ as inputs.
Then we extract fine-level corresponding features using coarse matching indices and flatten them to 1-D vectors $F^{A}$ , $F^{B}$.
\\\noindent\textbf{Bidirectional Refinement.}
We consider the central pixel of grids as keypoints $P^{A}$, $P^{B}$ with descriptors $F^{A}$ , $F^{B}$.
As show in \cref{fig:grid},
we define the grid center $P_{q}^{A}$ as the query point,
and its coarse corresponding keypoint on the reference image grid is $P_{q}^{B}$.
Due to quantization errors during supervision,
for the query points $P_{q}^{A}$,
there is an offset between the ground truth keypoint $P_{r}^{A}$ and the coarse corresponding keypoint $P_{q}^{B}$.
Similarly, we found that using point $P_{q}^{B}$ as the query point is dual.
So we propose a bidirectional refinement strategy to obtain double fine matches with a single slight inference.
We concatinate $F^{A}$ , $F^{B}$ in sequence as query features $F_{q}^{A}$ , $F_{q}^{B}$,
and reference features $F_{r}^{B}$ , $F_{r}^{A}$ in the reverse order.
Then, they are passed through their respective query and reference encoders,
each consisting of a lightweight Multi-Layer Perceptron (MLP).
Subsequently, the corresponding features are concatenated along the descriptive dimension and then merged through another MLP.
\\\noindent\textbf{Axis-Based Regression Head.}
Inspired by \cite{li2022simcc, li2021human, nibali2018numerical},
regressing numerical coordinates directly from latent features is extremely fast, yet it lacks spatial generalization and robustness.
To facilitate model learning,
we design a lightweight Axis-Based Regression Head (ABRHead) with Soft Coordinate Classification (SCC) as shown in the bottom-right of the \cref{fig:network}.
Taking the X-axis as an example,
the merged feature first passed through linear layers to reduce the number of output dimension to $N$+1.
The $N$-D tensor is passed through soft-argmax \cite{nibali2018numerical} to predict a location parameter $\mu$,
which indicates the index of the maximum response in continuous coordinate space from a classification view.
The another 1-D tensor is passed through a sigmoid activation function to predict a scale parameter $\mathcal{\sigma}$.
The output $\mu$ and $\mathcal{\sigma}$ are used to shift and scale the distribution generated by a flow model \cite{dinh2016density}, respectively.
SCC, which utilizes $N$ bins, implicitly encodes local coordinate information on the one hand, thereby reducing the learning difficulty.
On the other hand, it avoids the issue of regression method values exceeding the local window boundary.

Besides, we use RLE Loss \cite{li2021human} to supervise the prediction results of the network (refer to \cref{sec:loss}).
The predict $\mu$ is equivalent to the normalized offset $\Delta $,
which represents the distance from the predicted keypoint coordinate to the center of the grid along the current axis.
Additionally, $\phi$ represents the parameters of the flow model,
which is not required during the inference process, thus avoiding additional overhead during testing.
\\\noindent\textbf{Fine Matching Selection.}
The scale parameter $\mathcal{\sigma}$ reflects the standard deviation of the predict distribution.
The model will output a larger $\mathcal{\sigma}$ for a more uncertain result.
Therefore, the prediction confidence can be obtained by:
\begin{equation}
	\mathcal{P}_{f} = 1-\frac{\sigma _{x}+\sigma _{y}}{2}
	\label{eq:fine_th}
\end{equation}
where $\sigma _{x}$ and $\sigma _{x}$ represent the $\sigma$ on X- and Y-axis respectively.
For each bidirectional matching pair, we keep the more confident one while requiring it to be above the fine-level threshold $\theta _{f}$ to enhance the matching precision.
\subsection{Loss Function}
\label{sec:loss}
\textbf{Coarse-Level Loss Function.}
To generate the coarse-level ground truth matches $\mathcal{M}_{c}$,
we warp the grid centroids from input image ${I}^{A}$ to ${I}^{B}$ using relative camera poses and depth maps
at $\frac{1}{8}$ scale following previous works \cite{sarlin2020superglue,sun2021loftr,wang2024eloftr}.
The matching probability matrix $\mathcal{P}_{c}$ produced by dual-softmax is supervised by minimizing the focal loss \cite{ross2017focal}:
\begin{equation}
	\mathcal{L}_{c} = - \frac{1}{\left | \mathcal{M}_{c}\right |}\displaystyle\sum_{\left \langle i,j\right \rangle \in \mathcal{M}_{c}}^{} \alpha \left ( 1- \mathcal{P}_{c}{\left \langle i,j\right \rangle}\right )^{\gamma}\log_{}{\mathcal{P}_{c}{\left \langle i,j\right \rangle}}
	\label{eq:coarse_loss}
\end{equation}
where $\alpha$ and $\gamma$ are respectively defined as weighting factor and focusing parameter.
\\\noindent\textbf{Fine-Level Loss Function.}
We employ the residual log-likelihood estimation (RLE) \cite{li2021human} loss to improve the offset regression performance,
which can be defined as follows:
\begin{equation}
	\mathcal{L}_{f} = -\log\mathcal{G}_{\phi}\left ( \hat{x}\right ) -\log\mathcal{Q}_{\phi}\left ( \hat{x}\right ) + \log\mathcal{\sigma}
	\label{eq:fine_loss}
\end{equation}
where $\mathcal{G}_{\phi}\left ( \hat{x}\right )$ is the distribution learned by the normalizing flow model ${\phi}$,
$\mathcal{Q}_{\phi}\left ( \hat{x}\right )$ is a simple Laplace distribution,
and $\mathcal{\sigma}$ is the prediction scale parameter.
Specifically, the Laplace distribution loss item about $\mathcal{Q}_{\phi}\left ( \hat{x}\right )$ can be defined as:
\begin{equation}
	\mathcal{Q}_{\phi}\left ( \hat{x}\right )= \displaystyle\sum_{\mathcal{M}_{f}}^{}\frac{1}{\sigma}e^{-\frac{\left | \mu^{gt} -\mu \right |}{2\sigma}}
	\label{eq:q_loss}
\end{equation}
where ${\mathcal{M}_{f}}$ is the ground truth fine-level matches,
which is a subset of correctly predicted coarse-level matches $\tilde{{\mathcal{M}_{c}}}$.
The $\mu^{gt}$ is the corresponding ground truth offsets.

The total loss is the weighted sum of coarse-level and fine-level matching loss as follows:
\begin{equation}
	\mathcal{L} = \lambda _{c}\mathcal{L}_{c} + \lambda _{f} \mathcal{L}_{f}
	\label{eq:total_loss}
\end{equation}
\subsection{Implementation Details}
The backbone feature widths from $\frac{1}{2}$ scale to $\frac{1}{32}$ scale are [32, 64, 128, 256, 256].
We set $L$ to 2 in the CIM to transform deep correlations.
The coordinate bins number $N$ in ABRHead is 16.
The attention scale factor $s$ is set to 20.
Following \cite{wang2024eloftr}, to demonstrate the generalization ability of EDM,
we trained it on the outdoor MegaDepth dataset and evaluated it on all tasks and datasets in our experiments.
During the training phase,
images are resized and padded to the size of 832$\times$832.
The training process utilizes the AdamW optimizer with an initial learning rate of 2$e$-3 and a batch size of 32 on 8 NVIDIA 3090 GPU.
The model converges in 6 hours, which is extremely fast compared to other methods.
For loss function,
the focal loss parameters $\alpha$ and $\gamma$ are set to 0.25 and 2 respectively.
Then we set $\lambda _{c}$ to 1 for coarse-level loss weight and $\lambda _{f}$ to 0.2 for fine-level loss weight.
The coarse-level threshold $\theta _{c}$ is usually set to 5$e$-2, while fine-level threshold $\theta _{f}$ is set to 1$e$-6.
\section{Experiments}
\label{sec:experiments}
\subsection{Relative Pose Estimation}
\noindent\textbf{Datasets.}
We follow the test settings of the previous methods \cite{sun2021loftr, wang2024eloftr, sarlin2020superglue},
selecting 1500 image pairs from the indoor ScanNet \cite{dai2017scannet} dataset and outdoor MegaDepth \cite{MegaDepthLi18} dataset, respectively.
For ScanNet, we resize all images to 640$\times$480 resolution.
For MegaDepth, images are resized to 832$\times$832 for training and 1152$\times$1152 for validation.
\\\noindent\textbf{Evaluation Protocol.}
Following SuperGlue (SG) \cite{sarlin2020superglue} and LoFTR \cite{sun2021loftr},
the relative pose error is defined as the maximum of angular errors in rotation and translation.
We report the area under the cumulative curve (AUC) of the relative pose error under multiple thresholds,
including $5^\circ$, $10^\circ$, and $20^\circ$.
In addition, the pairwise matching runtime on the ScanNet dataset is reported to explain the accuracy-efficiency tradeoffs.
Specifically, a single NVIDIA 3090 GPU is used to measure the latency of all methods.
\\\noindent\textbf{Results.}
As shown in \cref{tab:megadepth_and_scannet},
EDM shows superior performance compared with sparse and semi-dense methods on both datasets,
with the exception of a slightly lower AUC@$10^{\circ}$ on the ScanNet dataset compared to ASpanFormer \cite{chen2022aspanformer}.
Specifically, our method surpasses the recent semi-dense baseline ELoFTR \cite{wang2024eloftr} on all metrics,
with a significant speed improvement.
\begin{table}
	\footnotesize
	\centering
	\begin{tabular}{llccc}
		\toprule
		\multirow{2}*{Category} & \multirow{2}*{Method}                                                & \multicolumn{3}{c}{Homography est. AUC}                                 
		\\\cmidrule(lr){3-5}
		                        &                                                                      & @3px                                    & @5px          & @10px
		\\\midrule
		\multirow{2}*{Sparse}
		                        & DISK \cite{tyszkiewicz2020disk} + NN                                 & 52.3                                    & 64.9          & 78.9          \\
		                        & SP \cite{detone2018superpoint} + SG \cite{sarlin2020superglue}       & 53.9                                    & 68.3          & 81.7          \\
		                        & SP \cite{detone2018superpoint} + LG \cite{lindenberger2023lightglue} & 54.2                                    & 68.3          & 81.5
		\\\midrule
		\multirow{7}*{Semi-Dense}
		                        & DRC-Net \cite{li2020dual}                                            & 50.6                                    & 56.2          & 68.3          \\
		                        & Patch2Pix \cite{zhou2021patch2pix}                                   & 59.3                                    & 70.6          & 81.2          \\
		                        & LoFTR \cite{sun2021loftr}                                            & 65.9                                    & 75.6          & 84.6          \\
		                        & TopicFM \cite{giang2023topicfm}                                      & 67.3                                    & 77.0          & 85.7          \\
		                        & ASpanFormer \cite{chen2022aspanformer}                               & 67.4                                    & 76.9          & 85.6          \\
		                        & EfficientLoFTR \cite{wang2024eloftr}                                 & 66.5                                    & 76.4          & 85.5          \\
		                        & Ours                                                                 & \textbf{68.5}                           & \textbf{78.1} & \textbf{86.6}
		\\\bottomrule
	\end{tabular}
	\caption{Homography estimation on HPatches.}
	\label{tab:hpatches}
\end{table}
\subsection{Homography Estimation}
\noindent\textbf{Dataset.}
We evaluate our method on the widely adopted HPatches dataset \cite{balntas2017hpatches} for homography estimation.
\\\noindent\textbf{Evaluation Protocol.}
Following \cite{sarlin2020superglue, sun2021loftr, giang2023topicfm},
we resize input images to 480px in the smallest dimension and select the top 1000 matches.
We compute the mean reprojection error for the four corners and report the AUC values under 3, 5, and 10-pixel thresholds.
For fairness, we use the same OpenCV RANSAC with identical parameters to estimate homography for all comparative methods.
\\\noindent\textbf{Results.}
As presented in \cref{tab:hpatches},
Our EDM notably outperforms other methods under all thresholds,
demonstrating its effectiveness for homography estimation.
\begin{table}
	\footnotesize
	\centering
	\begin{tabular}{lcc}
		\toprule
		\multirow{2}*{Method}                                                & \multicolumn{1}{c}{DUC1}                                                           & \multicolumn{1}{c}{DUC2}             \\
		\cmidrule(lr){2-3}
		                                                                     & \multicolumn{2}{c}{(0.25m,$2^{\circ}$) / (0.5m,$5^{\circ}$) / (1.0m,$10^{\circ}$)}
		\\\midrule
		SP \cite{detone2018superpoint} + SG \cite{sarlin2020superglue}       & 47.0 / 69.2 / 79.8                                                                 & 53.4 / 77.1 / 80.9                   \\
		SP \cite{detone2018superpoint} + LG \cite{lindenberger2023lightglue} & 49.0 / 68.2 / 79.3                                                                 & 55.0 / 74.8 / 79.4                   \\
		LoFTR \cite{sun2021loftr}                                            & 47.5 / 72.2 / 84.8                                                                 & 54.2 / 74.8 / 85.5                   \\
		TopicFM \cite{giang2023topicfm}                                      & 52.0 / \textbf{74.7} / \textbf{87.4}                                               & 53.4 / 74.8 / 83.2                   \\
		ASpanFormer \cite{chen2022aspanformer}                               & 51.5 /73.7 / 86.0                                                                  & 55.0 / 74.0 / 81.7                   \\
		PATS \cite{ni2023pats}                                               & \textbf{55.6} / 71.2 / 81.0                                                        & 58.8 / 80.9 / 85.5                   \\
		EfficientLoFTR \cite{wang2024eloftr}                                 & 52.0 / \textbf{74.7} / 86.9                                                        & 58.0 / 80.9 / \textbf{89.3}          \\
		Ours                                                                 & 51.5 / 72.7 / 85.9                                                                 & \textbf{59.5} / \textbf{82.4} / 88.5
		\\\bottomrule
	\end{tabular}
	\caption{Results of visual localization on InLoc dataset.}
	\label{tab:inloc}
\end{table}
\begin{table}
	\footnotesize
	\centering
	\begin{tabular}{lcc}
		\toprule
		\multirow{2}*{Method}                                                & \multicolumn{1}{c}{Day}                                                            & \multicolumn{1}{c}{Night}    \\
		\cmidrule(lr){2-3}
		                                                                     & \multicolumn{2}{c}{(0.25m,$2^{\circ}$) / (0.5m,$5^{\circ}$) / (1.0m,$10^{\circ}$)}
		\\\midrule
		SP \cite{detone2018superpoint} + SG \cite{sarlin2020superglue}       & 89.8 / 96.1 / \textbf{99.4}                                                        & 77.0 / 90.6 / \textbf{100.0} \\
		SP \cite{detone2018superpoint} + LG \cite{lindenberger2023lightglue} & \textbf{90.2} / 96.0 / \textbf{99.4}                                               & 77.0 / 91.1 / \textbf{100.0} \\
		LoFTR \cite{sun2021loftr}                                            & 88.7 / 95.6 / 99.0                                                                 & \textbf{78.5} / 90.6 / 99.0  \\
		TopicFM \cite{giang2023topicfm}                                      & \textbf{90.2} / 95.9 / 98.9                                                        & 77.5 / 91.1 / 99.5           \\
		ASpanFormer \cite{chen2022aspanformer}                               & 89.4 / 95.6 / 99.0                                                                 & 77.5 / 91.6 / 99.5           \\
		PATS \cite{ni2023pats}                                               & 89.6 / 95.8 / 99.3                                                                 & 73.8 / \textbf{92.1} / 99.5  \\
		EfficientLoFTR \cite{wang2024eloftr}                                 & 89.6 / \textbf{96.2} / 99.0                                                        & 77.0 / 91.1 / 99.5           \\
		Ours                                                                 & 89.1 / \textbf{96.2} / 98.8                                                        & 77.0 / \textbf{92.1} / 99.5
		\\\bottomrule
	\end{tabular}
	\caption{Results of visual localization on Aachen v1.1 dataset.}
	\label{tab:aachen}
\end{table}
\subsection{Visual Localization}
\noindent\textbf{Datasets and Evaluation Protocols.}
Following \cite{sarlin2020superglue, sun2021loftr},
We assess our method on the InLoc \cite{taira2018inloc} dataset and Aachen v1.1 \cite{sattler2018benchmarking} dataset for visual localization,
within the open-sourced localization pipeline HLoc \cite{sarlin2019coarse}.
\\\noindent\textbf{Results.}
As shown in \cref{tab:inloc} and \cref{tab:aachen},
EDM performs comparably to sparse and semi-dense methods on the InLoc dataset and Aachen v1.1 dataset,
demonstrating robust generalization in visual localization.

\begin{figure}[t]
	\centering
	\includegraphics[width=1\linewidth]{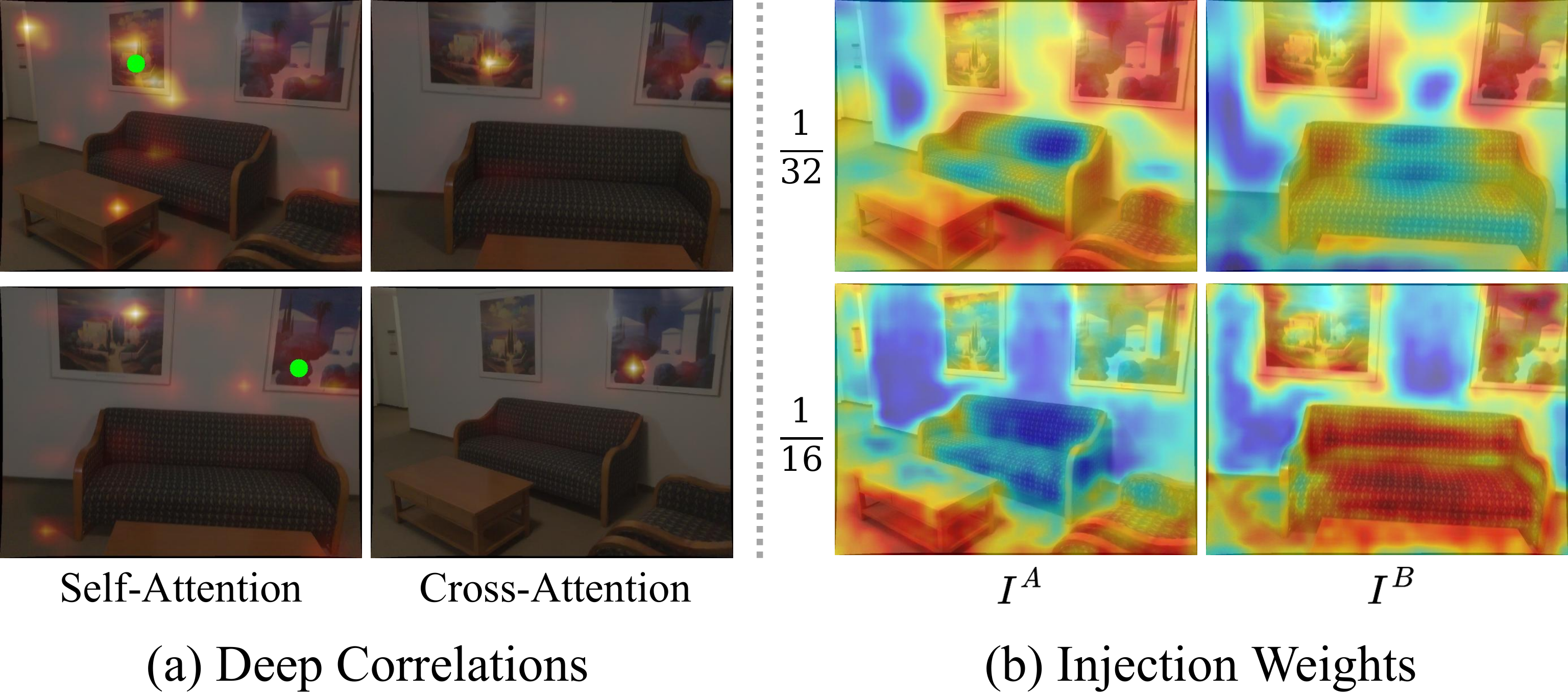}
	\caption{
		\textbf{Attention Visualization.} (a) Deep correlations.The green dots represent the query points.
		(b) Injection weights. Significant response values usually located in detail-rich regions.
	}
	\label{fig:attention}
\end{figure}
\begin{figure*}
	\centering
	\includegraphics[width=1\linewidth]{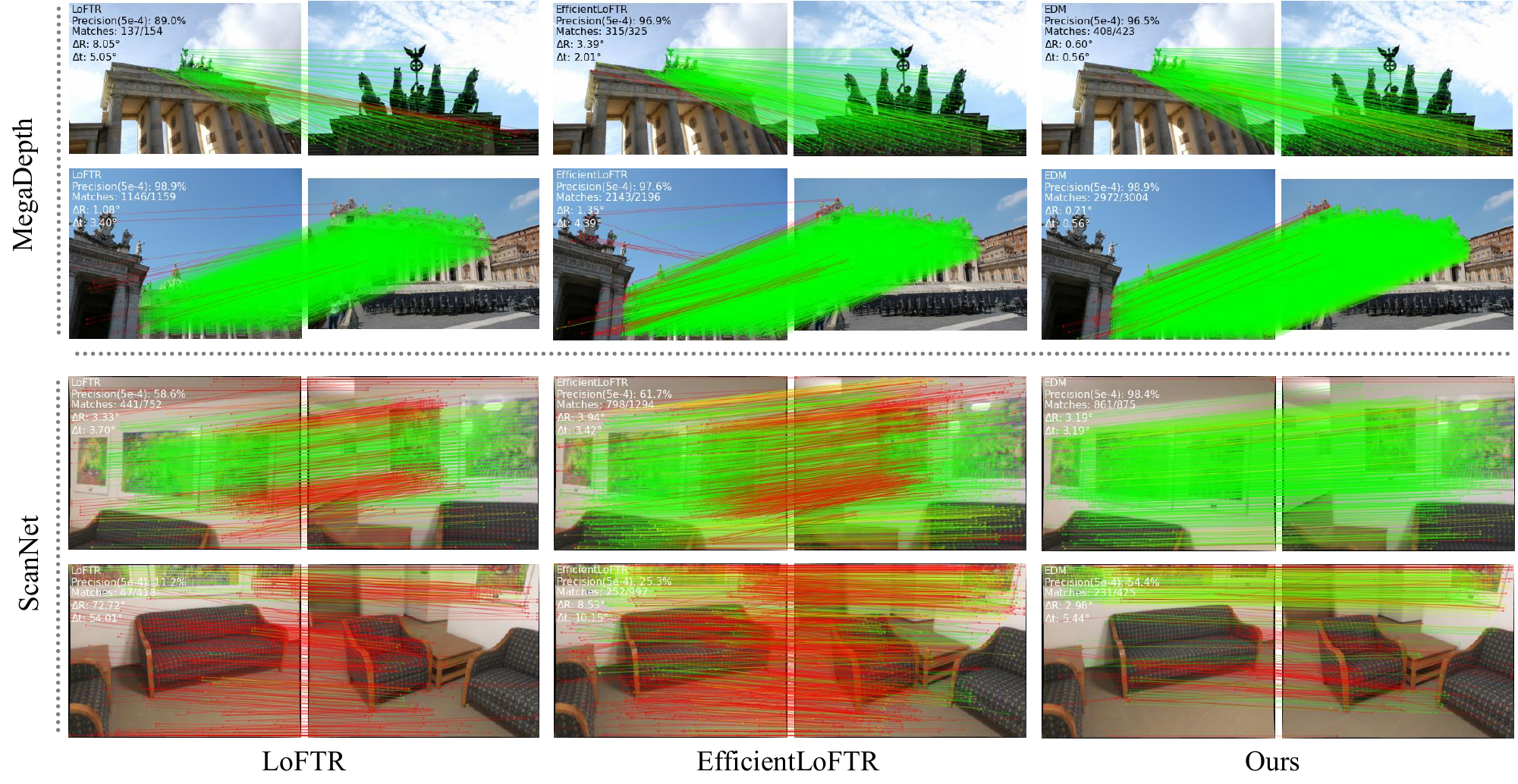}
	\caption{\textbf{Qualitative Comparisons.}
		Compared with LoFTR \cite{sun2021loftr} and EfficientLoFTR \cite{wang2024eloftr}, our method is more robust in scenarios with large viewpoint changes and repetitive semantics.
		The red color indicates epipolar error beyond 5$e$-4 in the normalized image coordinates.
	}
	\label{fig:qualitative}
\end{figure*}
\subsection{Understanding EDM}
\noindent\textbf{Weight Analysis.}
In the CIM,
we employ self- and cross-attention alternately at deeper layers to capture feature correlations enriched with high-level context information, such as semantics and structures.
To explain this process, we selected several query points and visualized the outcomes of self- and cross-attention separately.
Specifically, we summed and normalized the weight maps corresponding to the same input image and the same type of attention,
upsampled and overlaid them on the original image.
As depicted in \cref{fig:attention} (a), in the context of self-attention, the larger response points are more dispersed across different semantic regions.
Conversely, in cross-attention, the significant response points are more concentrated in proximity to the potential matching points.

In the ILs after modeling feature correlations,
the low-resolution global features, characterized by a larger receptive field and rich context information,
are fed into a CBA block to generate weights that determine the level of detail retention for the local features.
As shown in \cref{fig:attention} (b),
we overlay two layers of weight maps onto the input images.
The weight maps at $\frac{1}{32}$ and $\frac{1}{16}$ scale layers exhibit different focuses,
but the more prominent response values generally cluster in regions with distinct details.
\\\noindent\textbf{Qualitative Results Visualization.}
As shown in \ref{fig:qualitative}.
Our approach is able to extract more adequate and accurate matches compared to LoFTR \cite{sun2021loftr} and ELoFTR \cite{wang2024eloftr},
even in challenging scenes characterized by wide viewpoints, repetitive patterns and textureless regions.
Previous methods primarily focused on low-level local features,
often struggle with strong repetitive structures in indoor environments, like similar paintings or sofas.
By leveraging CIM, EDM correlates high-level context information across views, thus enhancing matching capability.
\\\noindent\textbf{Image Size Analysis.}
As shown in \cref{tab:image_size},
we evaluate the performance and efficiency variations of our method and ELoFTR \cite{wang2024eloftr} across different image sizes.
Employing a larger image size leads to an accuracy enhancement, albeit at the expense of a slower speed.
Our method significantly outperforms ELoFTR \cite{wang2024eloftr} at all resolutions under both Mixed-Precision and FP32 configurations.
\begin{table}
	\fontsize{6pt}{8pt}\selectfont
	\centering
	\begin{tabular}{lccl}
		\toprule
		\multirow{2}*{Resolution}       & \multirow{2}*{Method}        & \multicolumn{1}{c}{Pose Est. AUC}        & \multicolumn{1}{c}{Runtime (ms)} \\
		\cmidrule(lr){3-3}
		\cmidrule(lr){4-4}
		                                &                              & @$5^{\circ}$/@$10^{\circ}$/@$20^{\circ}$ & Mixed-Precision / FP32           \\
		\midrule
		\multirow{2}*{640$\times$640}   & ELoFTR \cite{wang2024eloftr} & 51.0/67.4/79.8                           & 46.6 / 52.1                      \\
		                                & Ours                         & 52.2/68.9/80.9                           & \textbf{23.0 / 23.8 (-54.3\%)}   \\
		\midrule
		\multirow{2}*{800$\times$800}   & ELoFTR \cite{wang2024eloftr} & 53.4/70.0/81.9                           & 63.0 / 75.7                      \\
		                                & Ours                         & 54.3/70.8/82.4                           & 30.7 / 34.7 (-54.2\%)            \\
		\midrule
		\multirow{2}*{960$\times$960}   & ELoFTR \cite{wang2024eloftr} & 54.7/70.7/82.4                           & 90.2 / 114.9                     \\
		                                & Ours                         & 55.6/71.4/82.8                           & 44.9 / 52.8 (-54.0\%)            \\
		\midrule
		\multirow{2}*{1152$\times$1152} & ELoFTR \cite{wang2024eloftr} & 56.4/72.2/83.5                           & 142.1 / 185.0                    \\
		                                & Ours                         & 57.5/\textbf{73.2}/\textbf{84.2}         & 72.3 / 86.0 (-53.5\%)            \\
		\midrule
		\multirow{2}*{1408$\times$1408} & ELoFTR \cite{wang2024eloftr} & 56.2/73.1/83.4                           & 257.4 / 327.8                    \\
		                                & Ours                         & \textbf{57.6}/\textbf{73.2}/84.1         & 136.4 / 162.7 (-50.4\%)          \\
		\bottomrule
	\end{tabular}
	\caption{Comparison of image size on the MegaDepth dataset.}
	\label{tab:image_size}
\end{table}
\\\noindent\textbf{Stage Analysis.}
\begin{table}
	\scriptsize
	\centering
	\begin{tabular}{lccc}
		\toprule
		\multirow{2}*{Stage}   & \multicolumn{3}{c}{Runtime (ms)}                                                           \\
		\cmidrule(lr){2-4}
		                       & LoFTR \cite{sun2021loftr}        & ELoFTR \cite{wang2024eloftr} & Ours                     \\
		\midrule
		(a) Feature Extraction & 28.01                            & 9.12                         & \textbf{4.00 (-56.1\%)}  \\
		(b) Feature Transform  & 17.77                            & 12.52                        & \textbf{8.20 (-34.5\%)}  \\
		(c) Coarse Matching    & 7.80                             & 7.71                         & \textbf{2.28 (-70.4\%)}  \\
		(d) Fine Matching      & 18.23                            & 9.67                         & \textbf{2.26 (-76.6\%)}  \\
		\midrule
		Overall                & 71.81                            & 39.02                        & \textbf{16.74 (-57.1\%)} \\
		\bottomrule
	\end{tabular}
	\caption{Runtime comparisons for each stage on ScanNet dataset.}
	\label{tab:stages_runtime}
\end{table}
We evaluated the running time of each stage of our method on the ScanNet dataset at 640$\times$480 resolution,
and benchmarked it against the leading semi-dense matchers, LoFTR \cite{sun2021loftr} and ELoFTR \cite{wang2024eloftr}.
As presented in \ref{tab:stages_runtime},
our method achieves higher efficiency in all matching stages.
Specifically, compared to ELoFTR \cite{wang2024eloftr}, our method reduces the time consumption by 56.1\% in feature extraction, 34.5\% in feature transformation, 70.4\% in coarse matching, and 76.6\% in fine matching.
Finally, in terms of overall time, it is 2.3 times faster than ELoFTR.
\begin{table}
	\scriptsize
	\centering
	\begin{tabular}{lcc}
		\toprule
		\multirow{2}*{Method}                                                         & \multicolumn{1}{c}{Pose est. AUC}        & \multirow{2}*{Time(ms)} \\
		\cmidrule(lr){2-2}
		                                                                              & @$5^{\circ}$/@$10^{\circ}$/@$20^{\circ}$                           \\
		\midrule

		Ours Full                                                                     & 57.5/73.2/84.2                           & 86.0                    \\
		\midrule
		\textbf{(a) Feature Extraction}                                                                                                                    \\
		~(1) shadow-wide design (to $\frac{1}{8}$ scale)                              & 56.9/72.6/83.5                           & 109.8                   \\
		\midrule
		\textbf{(b) Feature Transform (CIM)}                                                                                                               \\
		~(2) replace QKNA with vanilla Attn.                                          & 56.7/72.9/84.0                           & 85.7                    \\
		~(3) $L$ = 0                                                                  & 51.8/67.3/78.8                           & 70.1                    \\
		~(4) $L$ = 1                                                                  & 55.9/71.5/82.9                           & 77.5                    \\
		~(5) $L$ = 4                                                                  & 57.6/73.2/84.1                           & 101.9                   \\
		~(6) replace ILs with element-wise sum                                        & 55.9/71.7/82.9                           & 85.0                    \\
		\midrule
		\textbf{(c) Coarse Matching}                                                                                                                       \\
		~(7) replace dual-softmax with ELoFTR's                                       & 57.3/73.0/84.0                           & 100.6                   \\
		~(8) replace coarse selection with MNN                                        & 57.2/72.7/83.9                           & 101.2                   \\
		~(9) negative log-likelihood loss                                             & 56.3/71.9/83.2                           & 86.0                    \\
		\midrule
		\textbf{(d) Fine Matching}                                                                                                                         \\
		~(10) replace fine matching with ELoFTR's                                     & 56.2/71.9/83.0                           & 113.8                   \\
		~(11) w/o bidirectional refinement                                            & 55.7/71.9/83.4                           & 84.0                    \\
		~(12) w/o fine selection by ${\sigma}$                                        & 57.0/72.7/83.8                           & 86.1                    \\
		~(13) replace $\mathcal{Q}_{\phi }\left ( \hat{x}\right )$ with Gussian dist. & 57.1/72.9/84.0                           & 86.0                    \\
		~(14) w/o Soft Coordinates Classification                                     & 56.5/72.2/83.4                           & 85.2                    \\
		~(15) replace RLE loss with L1 loss                                           & 53.9/70.6/82.4                           & 86.7                    \\
		~(16) replace RLE loss with L2 loss                                           & 53.6/70.1/82.2                           & 86.7                    \\
		\bottomrule
	\end{tabular}
	\caption{Ablation studies on the MegaDepth dataset at all steps, with average running times measured at 1152$\times$1152 resolution.}
	\label{tab:ablation}
\end{table}
\\\noindent\textbf{Ablation Study.}
For a comprehensive understanding of our method,
we conduct ablation studies at different stages on the MegaDepth dataset.
The results are shown in \cref{{tab:ablation}}.
(a) Feature Extraction.
(1) Following ELoFTR's shadow-wide network design results in decreased matching accuracy and a substantial increase in running time.
(b) Feature Transform.
(2) Adopting QKNA can improve evaluation metrics, especially for AUC@$5^{\circ}$.
(3-5) Setting $L$ = 2 achieves an optimal balance between performance and efficiency.
(6) Replacing ILs with a naive element-wise sum for multi-scale feature integration leads to a substantial performance drop.
(c) Coarse Matching.
(7) Our implementation of dual-softmax saves significant inference time compared to previous methods.
(8) Compared to MNN, our coarse matching selection strategy offers higher efficiency and precision.
(9) Focal loss improves performance compared to the negative log-likelihood loss in coarse matching supervised learning.
(d) Fine Matching.
(10) Replacing the entire stage with a high-resolution implementation of ELoFTR, leading to notable time overhead and a decline in performance.
(11) Bidirectional refinement yields significant performance gains with only a minor increase in time cost.
(12) Using ${\sigma}$ to select bidirectional fine matches for retaining more confident results can innocuously boost matching accuracy.
(13) Laplace distribution is a more suitable initial distribution for local feature matching than Gaussian distribution.
(14) SCC simplifies fine matching local offset regression.
(15-16) Compared to supervising regression results with L1 or L2 loss, the RLE loss significantly enhances regression accuracy without additional inference overhead.
\\\noindent\textbf{Limitations.}
EDM's significant relative efficiency advantage declines moderately with increasing image resolution due to deeper feature extraction layers.
However, semi-dense matchers generally achieve optimal performance without requiring extremely high resolutions.

\section{Conclusions}
\label{sec:conclusions}
Depart from the prevailing shallow-wide network design paradigm,
this paper introduces EDM,
an efficient deep feature matching network.
By alternately applying self- and cross-attention on low-resolution deep layers to model feature correlations,
and integrating global and local features through progressive correlation injection,
the proposed CIM notably reduces the number of tokens while capturing enriched high-level contextual information,
thereby enhancing both the matching accuracy and efficiency.
Besides, we design a novel lightweight bidirectional axis-based regression head to implicitly refine the coarse matches by estimating local coordinate offsets.
We also propose deployment-friendly matching selection strategies to filter accurate matches effectively at both coarse and fine matching stages.
As a result, 
EDM attains competitive performance in multiple benchmarks with superb efficiency by redesigning all the steps of the mainstream semi-dense matching pipeline,
opening up new prospects for time-sensitive image matching applications.
{
	\small
	\bibliographystyle{ieeenat_fullname}
	\bibliography{main}
}
\clearpage
\setcounter{page}{1}
\maketitlesupplementary

\section{Implementation Details}
\subsection{Training Details}
\begin{figure}[t]
	\centering
	\includegraphics[width=0.8\linewidth]{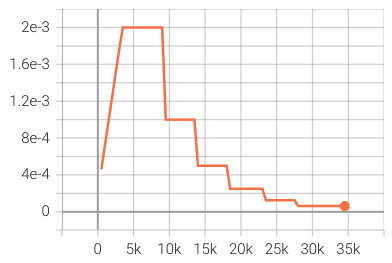}
	\caption{
		Learning rate curve over iterations.
	}
	\label{fig:lr}
\end{figure}

We adopt a multi-step training strategy to train our EDM model on the MegaDepth dataset for a total of 30 epochs.
The training begins with an initial learning rate of 2$e$-3, which undergoes a linear warmup phase lasting 3 epochs.
Following this, the learning rate is reduced by half every 4 epochs, starting from the eighth epoch.
The learning rate curve is depicted in \cref{fig:lr}.

Additionally, the supervision of fine matching relies on the predictions derived from coarse matching.
However, in the early training stages, the accuracy of these coarse matching predictions may be unsatisfactory.
In order to avoid distributed data parallel deadlocks and enhance the supervision of fine matching,
we pad the training samples with an additional 32 ground truth coarse matches for those with inadequate coarse matching predictions.

\section{More Experiments}
\subsection{CNN Backbone}
We utilize a simple variant of ResNet18 \cite{he2016deep} as our backbone, which achieves a minimum resolution of $\frac{1}{32}$.
To investigate the impact of various channel configurations, we conducted experiments on the MegaDepth dataset, as shown in \cref{tab:backbone}.
The final channel configuration we selected, $[32, 64, 128, 256, 256]$,
offers an optimal balance between efficiency and performance.

\subsection{Fine Matching Selection}
\begin{figure}[t]
	\centering
	\includegraphics[width=1\linewidth]{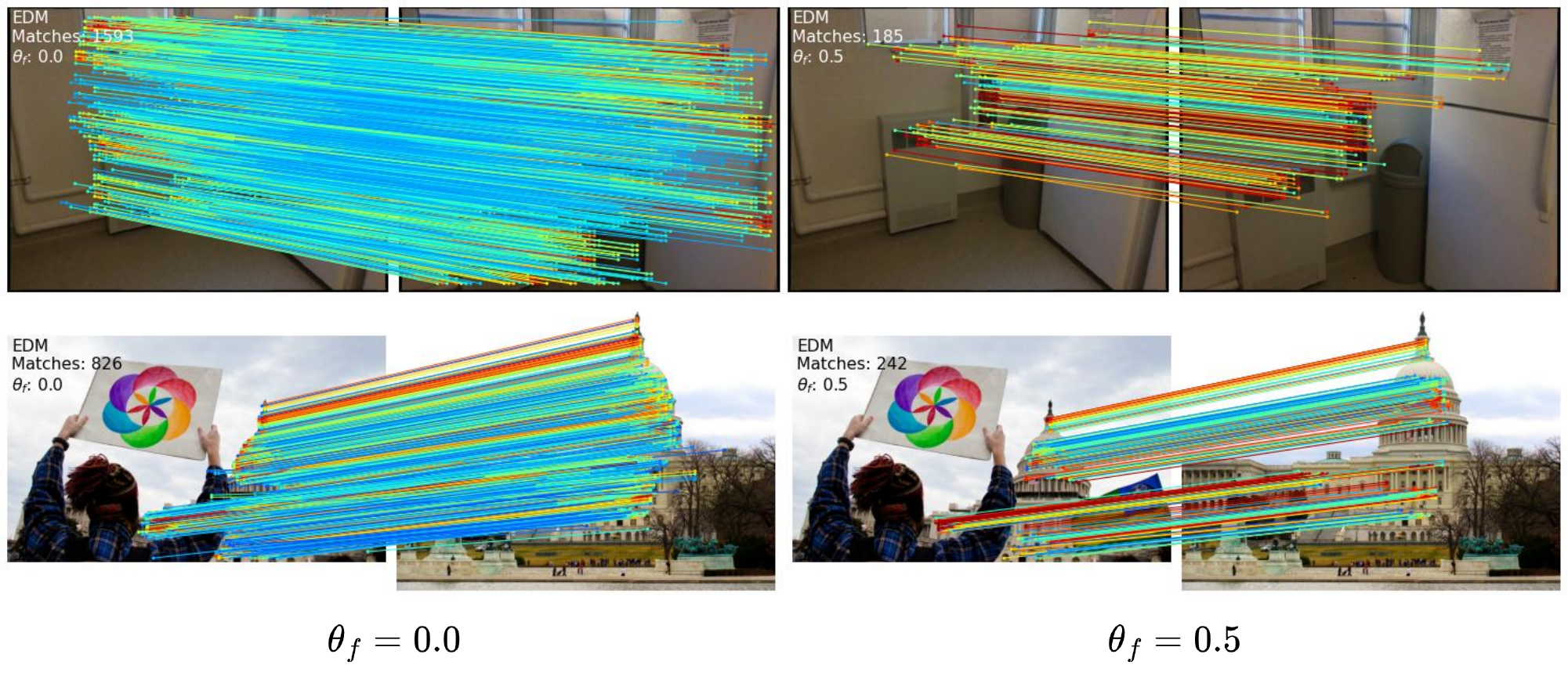}
	\caption{
		Impact of $\theta_{f}$ on fine-level matching filtering.
		Matches in areas with obvious details tend to have smaller $\sigma$, indicating higher confidence in these results.
	}
	\label{fig:theta_f}
\end{figure}

As presented in \cref{eq:fine_loss},
during the training process,
on the one hand,
it is desirable for the fine-level loss function $\mathcal{L}_{f}$ to constrain the value of $\sigma$ in term $\log\mathcal{\sigma}$ to be as small as possible.
On the other hand, in term $\log\mathcal{Q}_{\phi}\left ( \hat{x}\right )$,
when the distance between the predicted mean $\mu$ and the ground truth mean $\mu^{gt}$ is large,
the standard deviation $\sigma$ of the predict distribution tends to increase in order to mitigate the overall loss.
As demonstrated in \cref{fig:theta_f}, the matches that exhibit higher confidence $\mathcal{P}_{f}$ (indicated by a smaller $\mathcal{\sigma}$) are frequently found in image regions that contain abundant local details.
This observation suggests that the model leverages these detailed areas to make more precise and confident predictions.
\begin{table}
	\footnotesize
	\centering
	\begin{tabular}{lcccc}
		\toprule
		\multirow{2}*{backbone channels} & \multicolumn{3}{c}{Pose Estimation AUC} & \multirow{2}*{Time (ms)}                         \\
		\cmidrule(lr){2-4}
		                                 & @$5^{\circ}$                            & @$10^{\circ}$            & @$20^{\circ}$         \\
		\midrule
		$[16, 32, 64, 128, 256]$         & 55.9                                    & 71.8                     & 83.1          & 77.6  \\
		$[32, 64, 128, 128, 256]$        & 57.3                                    & 73.0                     & 84.0          & 84.2  \\
		$[32, 64, 128, 256, 256]$        & 57.5                                    & 73.2                     & 84.2          & 86.0  \\
		$[32, 64, 128, 256, 512]$        & 57.8                                    & 73.1                     & 84.0          & 96.0  \\
		$[ 64, 128, 256, 256, 256]$      & 58.0                                    & 73.5                     & 84.3          & 109.3 \\
		\bottomrule
	\end{tabular}
	\caption{The results of varying backbone channel numbers, from $\frac{1}{2}$ to $\frac{1}{32}$ scale, on the MegaDepth dataset.}
	\label{tab:backbone}
\end{table}
\begin{figure}[t]
	\centering
	\includegraphics[width=0.8\linewidth]{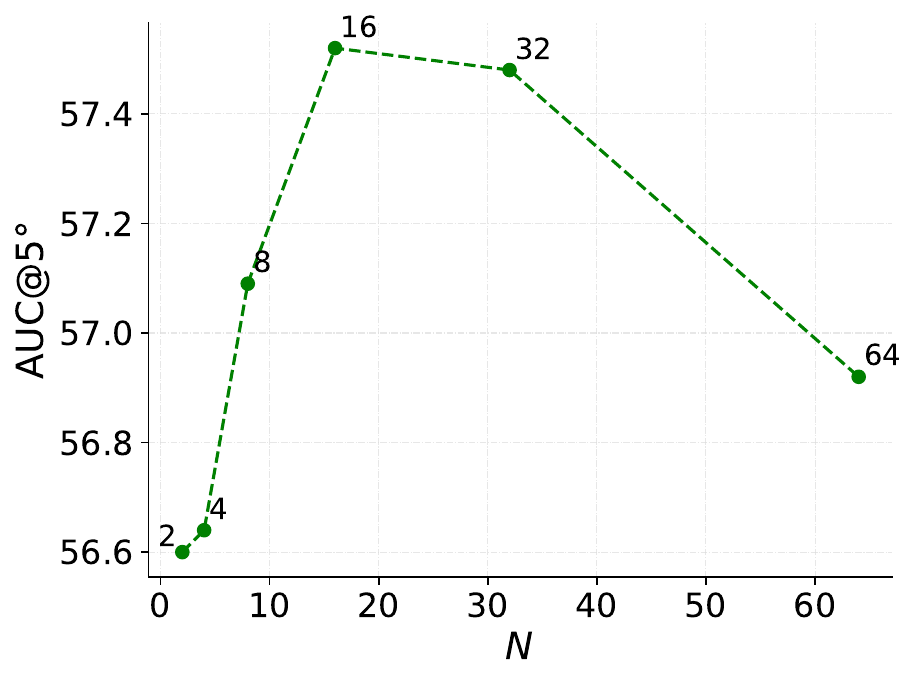}
	\caption{
		The impact of $N$ on matching accuracy.
	}
	\label{fig:scc_bins}
\end{figure}
\begin{table}
	\small
	\centering
	\begin{tabular}{lcccc}
		\toprule
		\multirow{2}*{$N$} & \multicolumn{3}{c}{Pose Estimation AUC}                                   \\
		\cmidrule(lr){2-4}
		                   & @$5^{\circ}$                            & @$10^{\circ}$  & @$20^{\circ}$  \\
		\midrule
		2                  & 56.60                                   & 72.34          & 83.56          \\
		4                  & 56.64                                   & 72.81          & 83.89          \\
		8                  & 57.09                                   & 72.75          & 83.75          \\
		16                 & \textbf{57.53}                          & \textbf{73.21} & \textbf{84.20} \\
		32                 & 57.48                                   & 73.02          & 84.19          \\
		64                 & 56.92                                   & 72.56          & 83.60          \\
		\bottomrule
	\end{tabular}
	\caption{The results of different $N$ on the MegaDepth dataset.}
	\label{tab:scc_bins_no}
\end{table}
\subsection{SCC Bins}
The Soft Coordinate Classification (SCC) bins number $N$ in the Axis-Based Regression Head (ABRHead) is set to 16 in our EDM.
Considering the X-axis as an illustrative example,
during the fine-level matching,
each 8$\times$8 grid along the X-axis is divided into $N$ bins,
so each pixel within this grid is mapped to $\frac{N}{8}$ bins.
The \cref{fig:scc_bins} demonstrates a trend in matching accuracy as $N$ varies.
Initially, as $N$ increases, the accuracy of matching improves, benefiting from the finer segmentation of pixels into bins.
However, as $N$ continues to grow, the complexity of the learning task also increases, leading to a gradual decline in matching accuracy.
Detailed experimental results supporting this observation are provided in \cref{tab:scc_bins_no}.
Additionally, the differences in network parameters caused by variation $N$ are relatively small,
rendering the comparison of efficiency between different settings unnecessary,
as they exhibit similar performance in terms of computational cost.

\subsection{TensorRT Runtime}
To further demonstrate the potential of our method in industrial applications,
we deployed the EDM with float32 precision based on ONNX Runtime with TensorRT engine,
and compared its runtime with that of the native PyTorch model.
The inference times for the same image pair on the identical hardware are presented in \cref{tab:tensorrt}.
With the acceleration provided by the TensorRT platform, our deployment-friendly model can achieve a higher efficiency.
Besides, due to the sensitivity of feature matching tasks to precision, we do not recommend reducing the numerical precision of a well-trained model.

\begin{table}
	\footnotesize
	\centering
	\begin{tabular}{lcc}
		\toprule
		Platform & Time (ms)     \\
		\midrule
		PyTorch & 16.74         \\
		TensorRT & \textbf{6.72} \\
		\bottomrule
	\end{tabular}
	\caption{Comparison of inference time on different platforms.
		The running times for an image pair with 640$\times$480 resolution are measured on a single NVIDIA 3090 GPU.}
	\label{tab:tensorrt}
\end{table}

\subsection{Batch Inference}
Our design prioritizes efficiency and deployment flexibility.
Our method enables data to be grouped into mini-batches for batch inference, 
thereby reducing average computational resource consumption overall.
As shown in \cref{tab:batch_infer},
inference latency measurements across batch sizes are benchmarked on a single NVIDIA 3090 GPU with 640$\times$480 resolution.
\begin{table}
	\footnotesize
	\centering
	\begin{tabular}{lcccccc}
		\toprule
		Batch Size   & 1    & 2    & 4    & 8    & 16    & 32    \\
		\midrule
		Runtime (ms) & 16.7 & 20.7 & 37.4 & 70.1 & 136.5 & 270.8 \\
		\bottomrule
	\end{tabular}
	\caption{Comparison of inference time on different batch size.}
	\label{tab:batch_infer}
\end{table}
\subsection{Other Efficiency Comparisons}
More efficiency comparisons in as \cref{tab:efficiency} shown,
the results further validate the effectiveness of our method.
\begin{table}
	\tiny
	\centering
	\begin{tabular}{lccccc}
		\toprule
		Method                                                               & Parameters(M) & FLOPs(G)      & Memory(MB)     & Runtime(ms)   & AUC@$5^{\circ}$ \\
		\midrule
		RoMa \cite{edstedt2024roma}                                          & 111.3         & 2586.3        & 2791.1         & 312.9         & \textbf{28.9}   \\
		\midrule
		SP \cite{detone2018superpoint} + LG \cite{lindenberger2023lightglue} & \textbf{8.9}  & 290.5         & 646.1          & 29.2          & 14.8            \\
		\midrule
		LoFTR \cite{sun2021loftr}                                            & 11.6          & 783.6         & 1029.6         & 71.8          & 16.9            \\
		ELoFTR \cite{wang2024eloftr}                                         & 15.1          & 420.9         & 985.3          & 39.0          & 19.2            \\
		EDM (ours)                                                           & 10.2          & \textbf{72.6} & \textbf{493.0} & \textbf{16.7} & 19.8            \\
		\bottomrule
	\end{tabular}
	\caption{More Efficiency Comparisons on ScanNet dataset.}
	\label{tab:efficiency}
\end{table}
\subsection{Additional Results on other RANSAC setting}
Recent semi-dense method JamMa \cite{lu2025jamma} introducing Mamba \cite{gu2023mamba} to enhance matching performance and efficiency,
employs more advanced poselib LO-RANSAC \cite{PoseLib} for evaluating relative pose estimation.
We follow the same setting as JamMa to further evaluate our method on the MegaDepth dataset.
Specifically, test images are resized and padded to 832$\times$832,
and the inlier pixel threshold of LO-RANSAC is set to 0.5.
The results are shown in \cref{tab:loransac},
our method outperforms all previous semi-dense methods in terms of accuracy and efficiency.
\begin{table}
	\scriptsize
	\centering
	\begin{tabular}{lcccc}
		\toprule
		\multirow{2}*{Method}        & \multicolumn{3}{c}{Pose Estimation AUC (LO-RANSAC)} & \multirow{2}*{Time (ms)}                                    \\
		\cmidrule(lr){2-4}
		                             & AUC@$5^{\circ}$               & AUC@$10^{\circ}$         & AUC@$20^{\circ}$ &
		\\\midrule
		LoFTR \cite{sun2021loftr}    & 62.1                          & 75.5                     & 84.9             & 134.8         \\
		ELoFTR \cite{wang2024eloftr} & 63.7                          & 77.0                     & 86.4             & 82.5          \\
		JamMa \cite{lu2025jamma}     & 64.1                          & 77.4                     & 86.5             & 84.3          \\
		Ours                         & \textbf{64.7}                 & \textbf{77.8}            & \textbf{86.8}    & \textbf{38.5} \\
		\bottomrule
	\end{tabular}
	\caption{Results of Relative Pose Estimation on MegaDepth Dataset following JamMa's setting.}
	\label{tab:loransac}
\end{table}

\section{More Visualizations}
\subsection{More Intuitive Explanation of Fine Matching}
\begin{figure}
	\centering
	\includegraphics[width=1\linewidth]{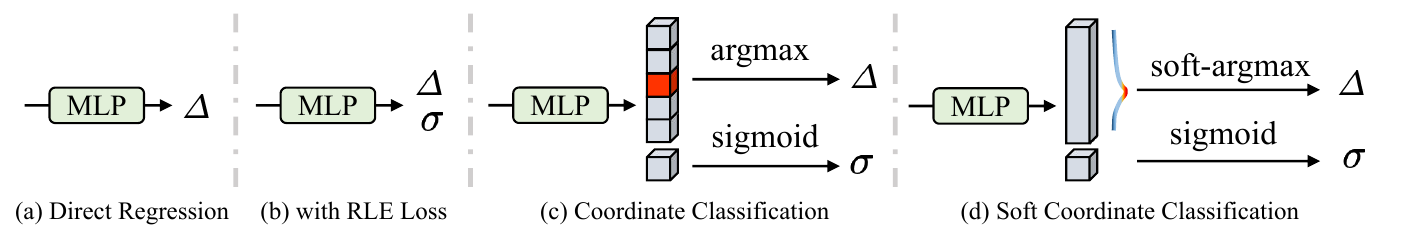}
	\caption{Regression Paradigms.}
	\label{fig:regression}
\end{figure}
\begin{figure}
	\centering
	\includegraphics[width=1\linewidth]{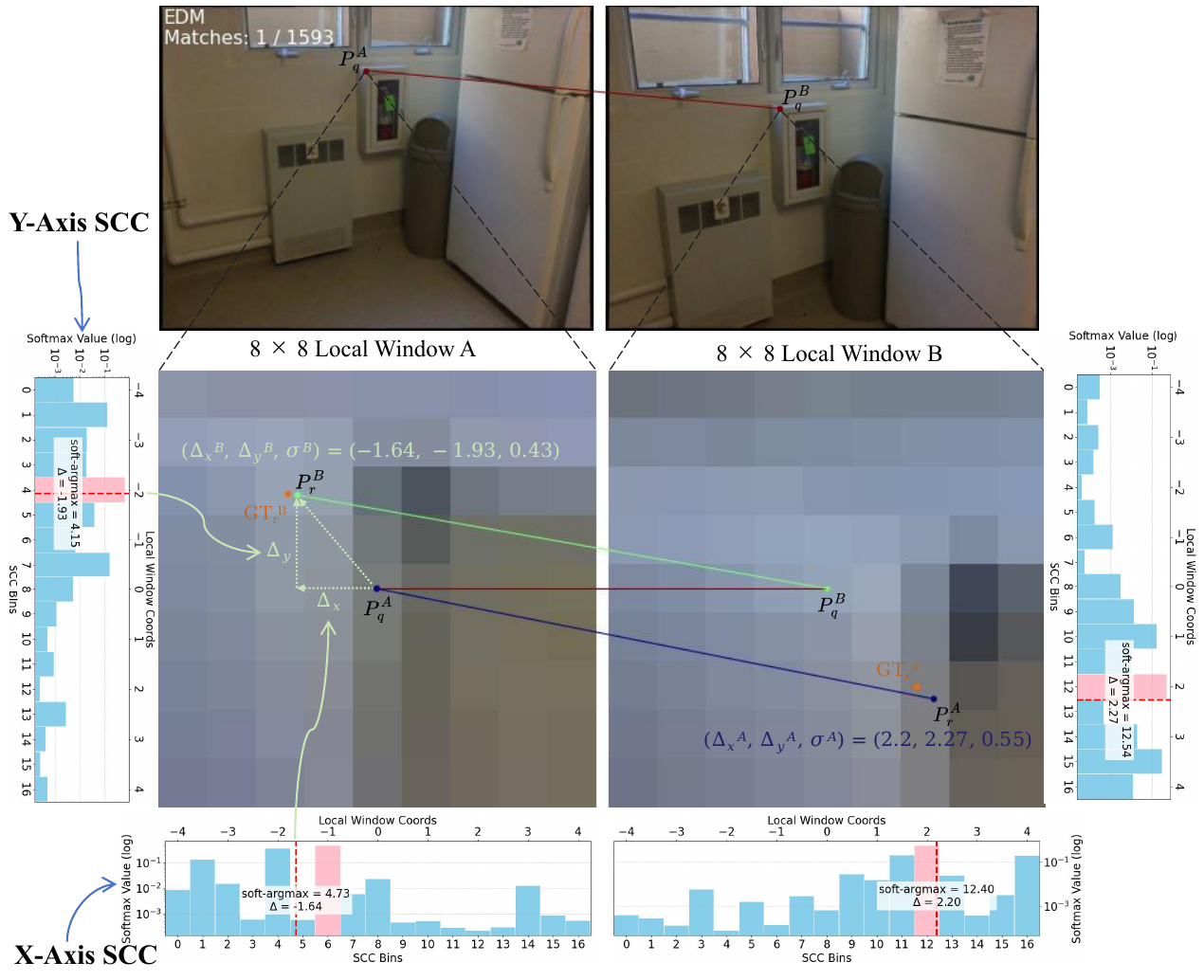}
	\caption{Explanation of Bidirectional Axis-Based Matching.}
	\label{fig:Bidirectional}
\end{figure}
We further explained our bidirectional axis-based matching and regression pipeline.
\cref{fig:regression} shows our thinking and improvement process regarding different regression paradigms in the task of implicit matching coordinate estimation.

Furthermore, as shown in \cref{fig:Bidirectional},
we illustrate a correctly predicted coarse match from real inference data by visualizing its corresponding 8$\times$8 image patch with both network predictions and ground truth overlaid.
This provides a realistic and intuitive explanation to highlight the differences from previous methods.
It can be observed that the bidirectional axis-based regression head operates as expected.
Specifically, it can be summarized as follows:
(a) Distinguishing the implicit encoding of local coordinates for axes reduces the optimization difficulty.
(b) Inherent bounding within the window eliminates regression outliers.
(c) Multimodal distribution facilitates the correction of imprecise maximum response values.
(d) In the two bidirectional fine matching of a pair of coarse correspondences, 
the one with lower standard deviation $\sigma$ (higher confidence score) typically has a smaller discrepancy with the ground truth, 
indicating higher accuracy.
\subsection{Failure Cases}
\begin{figure}
	\centering
	\includegraphics[width=1\linewidth]{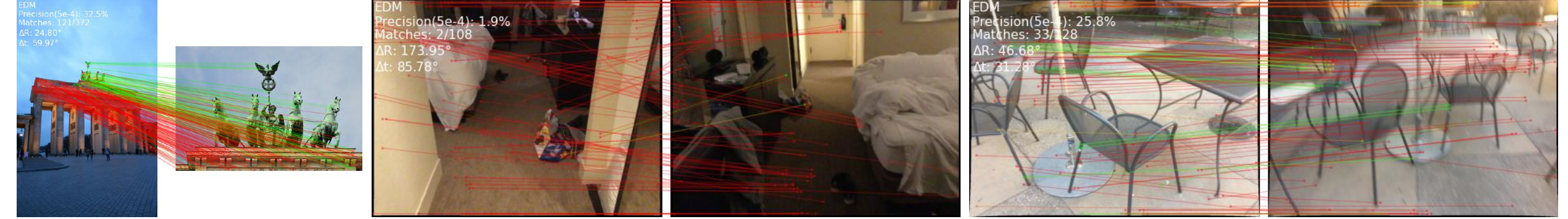}
	\caption{Failure Cases.}
	\label{fig:failure}
\end{figure}

As shown in \cref{fig:failure},
EDM's failure cases typically occur in scenarios with extreme scale and viewpoint variations or textureless regions.
\section{Future Work}
Although we have made improvements to each step of the detector-free matching pipeline,
the efficiency improvement of feature transformation is the least significant.
Even though the number of tokens has been significantly reduced,
the efficiency issue still persists due to the inherent characteristics of the Transformer.
In future work, we consider experimenting and replacing some components in our pipeline,
including more efficient backbones and feature transform mechanisms to further improve the accuracy and speed.

\end{document}